\documentclass{article}

\PassOptionsToPackage{numbers, compress}{natbib}
\usepackage[final]{neurips_2025}

\usepackage{enumitem}
\usepackage{amsmath}
\usepackage{amssymb}
\usepackage{graphicx}
\usepackage[utf8]{inputenc} % allow utf-8 input
\usepackage[T1]{fontenc}    % use 8-bit T1 fonts
\usepackage{hyperref}       % hyperlinks
\usepackage{url}            % simple URL typesetting
\usepackage{booktabs}       % professional-quality tables
\usepackage{amsfonts}       % blackboard math symbols
\usepackage{nicefrac}       % compact symbols for 1/2, etc.
\usepackage{microtype}      % microtypography
\usepackage{xcolor}         % colors
\usepackage{colortbl}
\usepackage{wrapfig}
\usepackage{subcaption}

\title{Top-Down Compression: Revisit Efficient Vision Token Projection for Visual Instruction Tuning}

\author{%
  Bonan Li\thanks{Equal Contribution} \\
    UCAS \& NUS\\
  \texttt{libonan@ucas.ac.cn} \\
  \And
  Zicheng Zhang$^{*}$ \\
  UCAS \\
  zhangzicheng19@mails.ucas.ac.cn \\
  \AND
  Songhua Liu \\
    NUS \\
  songhua.liu@u.nus.edu \\
  \And
  Weihao Yu \\
  NUS \\
  weihaoyu@u.nus.edu \\
  \And
  Xinchao Wang\thanks{Corresponding Author}\\
  NUS \\
  xinchao@nus.edu.sg \\
}

\begin{document}

\maketitle
\begin{abstract}
Visual instruction tuning aims to enable large language models to comprehend the visual world, with a pivotal challenge lying in establishing an effective vision-to-language projection. However, existing methods often grapple with the intractable trade-off between accuracy and efficiency. In this paper, we present \textbf{LLaVA-Meteor}, a novel approach designed to break this deadlock, equipped with a novel Top-Down Compression paradigm that strategically compresses visual tokens without compromising core information. Specifically, we construct a trainable \textit{Flash Global Fusion} module based on efficient selective state space operators, which aligns the feature space while enabling each token to perceive holistic visual context and instruction preference  at low cost. Furthermore, a local-to-single scanning manner is employed to effectively capture local dependencies, thereby enhancing the model's capability in vision modeling. To alleviate computational overhead, we explore a \textit{Visual-Native Selection} mechanism that independently assesses token significance by both the visual and native experts, followed by aggregation to retain the most critical subset. Extensive experiments show that our approach reduces visual tokens by 75\%–95\% while achieving comparable or superior performance across 12 benchmarks, significantly improving efficiency.
\end{abstract}

\section{Introduction}

Visual instruction tuning~\cite{liu2023visual,liu2024improved,li2022blip,li2023blip} has emerged as a promising pipeline to extend large language models (LLMs) into the visual domain, achieving notable progress and demonstrating impressive performance across a wide range of vision-language tasks~\cite{lei2024ez,wang2024rl,kim2024vlm,arefeen2024vita,ma2025does,jin2025logicad}, \textit{e.g.}, image captioning~\cite{luu2024questioning,wang2024cogvlm,zhang2024good}, visual question answering~\cite{khademi2023mm,wang2023fashionvqa} and open-vocabulary object detection~\cite{jinllms,wang2024marvelovd}. To perceive the visual world, current mainstream approaches (\textit{e.g.}, LLaVA~\cite{liu2023visual}) typically project sequential visual tokens into the linguistic space, which are subsequently incorporated with textual representation and jointly interpreted by the LLMs decoder. However, the sheer number of visual tokens significantly increases the computational burden on LLMs. For instance, the widely used CLIP ViT-L/336px~\cite{radford2021learning} encodes a single $\text{672} \times \text{1008}$ image into $\text{48} \times \text{72} = \text{3456}$ tokens.

To mitigate this, a variety of token compression techniques~\cite{koner2024lookupvit,jie2024token,song2025less,yang2024pvc,yang2024visionzip,zhu2024focusllava,choudhury2024don,shang2024llava} have been proposed to condense visual information into a reduced set of tokens, thereby enhancing model efficiency. Broadly, these approaches fall into two categories: Fusion and Selection. Fusion-based methods either apply predefined merging operations (\textit{e.g.}, PixelShuffle~\cite{li2024llava,chen2024internvl,dong2024internlm}), which often compromise the structural integrity of the features, or learn to condense the full token sequence into a smaller fixed set via trainable modules (\textit{e.g.}, Q-Former~\cite{bai2023qwen,li2023blip,cha2024honeybee,chu2024mobilevlm}), which typically require substantial additional parameters and large-scale datasets for effective training. In contrast, Selection-based approaches identify a subset of important tokens using task-agnostic visual cues (\textit{e.g.}, class token attention~\cite{yang2024visionzip,liu2024multi,zhang2024cls}), but often overlook instruction relevance, limiting their alignment with downstream tasks.

\begin{figure}
  \centering
  \resizebox*{1\linewidth}{!}{\includegraphics[width=\columnwidth]{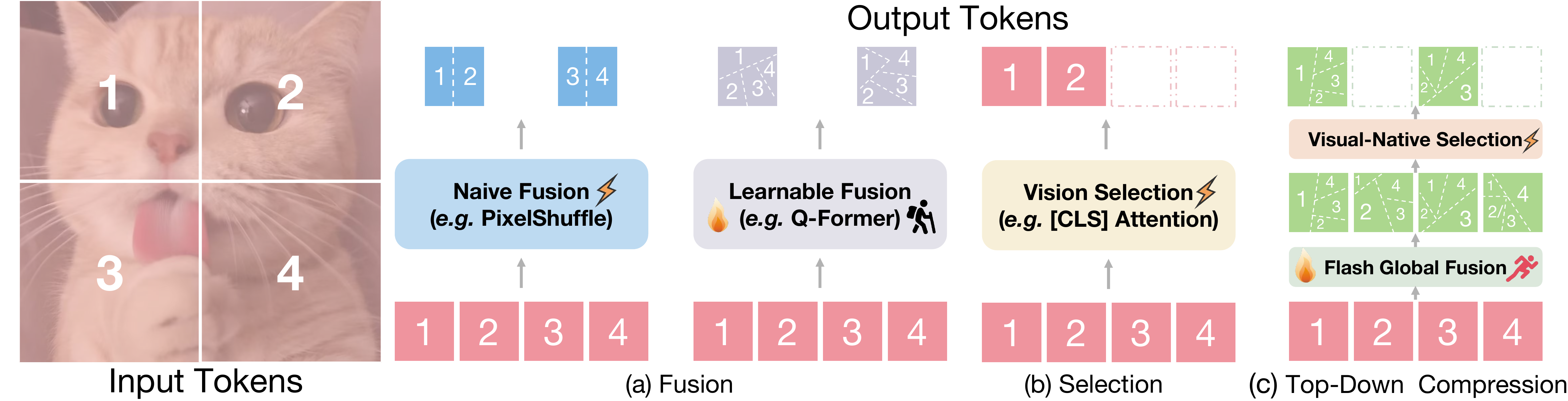}}
  \caption{Illustration of visual token compression strategies. (a) Naive and learnable fusion methods reduce token count via predefined rules or learned aggregation, but may lose structural fidelity or require heavy training. (b) Vision-based selection filters important tokens using attention, yet ignores instruction relevance. (c) Our Top-Down Compression combines Flash Global Fusion for global context propagation and instruction preference summarization, with Visual-Native Selection to jointly evaluate token importance from visual and native perspectives.}
  \label{fig:motivation}
\end{figure}

Despite notable progress, achieving the best of both worlds—accuracy and efficiency—remains an open challenge in visual instruction tuning. Existing fusion-based compression methods typically reduce the number of tokens either by predefined merging rules or through learned token aggregation. While effective in shortening the input length, such approaches often disrupt the structural integrity of visual features or introduce significant training complexity. To address this, we introduce a novel fusion module that retains the full token sequence and enhances each token's representation through global information propagation at low cost. Additionally, we design a dedicated instruction token that implicitly encodes instruction-level preferences by attending to the full visual context during global fusion. The instruction token is not conditioned on explicit prompts; rather, it learns to focus on instruction-relevant regions via weak supervision over large-scale vision-language data, where common user attention patterns such as phone number, faces, or salient objects are consistently reinforced. On the other hand, selection-based methods aim to identify the most important tokens using task-agnostic cues, such as attention scores from the class token. However, these approaches often overlook instruction relevance, leading to suboptimal token retention for instruction-following tasks. To mitigate this issue, we propose a native expert based on the instruction token, which complements the visual expert (class token) by providing instruction task guidance for token importance scoring. Integrating these two complementary components, we propose a unified visual token compression framework, termed Top-Down Compression. This paradigm first performs top-level global fusion to establish semantic context and task-aware priors, and subsequently applies bottom-level token selection guided by both visual and instruction-informed experts. As illustrated in Figure~\ref{fig:motivation}, the fusion stage (Top)  facilitates holistic context exchange and distills instruction priors into the instruction token. In the subsequent selection stage (Down), each token is evaluated from two perspectives: a visual expert that highlights globally salient regions, and a native expert that captures task-specific relevance. Based on these complementary signals, we select a compact and informative token subset to be forwarded to the LLMs, thereby achieving an effective balance between semantic richness and computational efficiency.

Building upon the proposed Top-Down Compression strategy, we present a novel and efficient vision-language model named \textbf{LLaVA-Meteor}, which comprises two key components: Flash Global Fusion (FGF) and Visual-Native Selection (VNS). In the FGF module, we leverage state space models (SSMs)~\cite{ssm} as the core operator due to their linear computational complexity and strong capability for long-range sequence modeling. This enables efficient propagation of global contextual information across the visual token sequence. To further enhance spatial understanding, we incorporate a local-to-single scanning strategy, which captures local dependencies by summarizing information from spatial neighborhoods before each token. In addition, a learnable instruction token is inserted at the center of the token sequence and trained to aggregate holistic semantic information while implicitly encoding instruction-related cues. In the VNS module, we adopt a dual-expert evaluation mechanism. The class token, obtained from a frozen vision encoder, serves as a visual expert and provides attention-based scores to reflect general visual importance. Meanwhile, the instruction token, enriched by the Flash Global Fusion module, serves as a native expert that evaluates instruction relevance by computing its similarity with each token in the sequence. To synthesize these two complementary perspectives, we aggregate their scores at each token index and select the Top-$K$ tokens with the highest combined importance as the final visual input to the LLMs.

In summary, our contributions can be summarized as follows: (\textit{$\romannumeral1$}) We construct the vision-language model LLaVA-Meteor upon the simple and effective Top-Down Compression paradigm, simultaneously achieving strong performance and high efficiency; (\textit{$\romannumeral2$}) We introduce a lightweight fusion module, Flash Global Fusion, which enables rapid global interaction and accurately distills instruction preferences; (\textit{$\romannumeral3$}) We propose an innovative hybrid selection strategy, Visual-Native Selection, which jointly assesses token importance through a visual and a native expert to guide the selection process; (\textit{$\romannumeral4$}) Extensive evaluations demonstrates competitive or superior performance across a wide range of benchmarks while significantly reducing the number of visual tokens. 

\section{Related Work}
\textbf{Visual instruction tuning} has become a cornerstone in extending large language models to understand and reason over visual content under textual instructions~\cite{hu2024bliva,wei2024vary,lin2023sphinx,lu2024deepseek,li2024monkey,team2024gemini,wei2024vary}. Early works such as Flamingo~\cite{alayrac2022flamingo} and BLIP-2~\cite{li2023blip} explored aligning vision-language representations through contrastive learning or image-caption matching objectives, demonstrating effectiveness in image-text retrieval and captioning. However, these models lacked flexible instruction-following capabilities.To bridge this gap, recent methods have introduced explicit multimodal instruction tuning. LLaVA~\cite{liu2023visual} established a lightweight yet effective pipeline by projecting CLIP-encoded image features into an LLMs (e.g., Vicuna) via a learnable MLP. This approach enabled strong instruction-following performance across multiple vision-language tasks and inspired variants like LLaVA-1.5~\cite{liu2024improved}, which incorporated high-resolution image handling and OCR-specific improvements. MiniGPT-4~\cite{zhuminigpt}, InstructBLIP~\cite{panagopoulou2023x}, Qwen-VL~\cite{bai2023qwen}, and CogVLM~\cite{wang2024cogvlm} further advanced this line by introducing multi-stage pretraining, hierarchical perception, or better instruction synthesis. Despite their success, these models are constrained by the large number of visual tokens injected into the LLMs, which significantly increases memory and latency. This has motivated a growing body of research on efficient visual token projection.

\textbf{Efficient vision token projection} aims to reduce the number of visual tokens passed into the language model while maintaining sufficient semantic fidelity for instruction understanding~\cite{gao2024tc,li2024llama,chen2024image,shang2024llava,li2024inference}. Existing approaches in this space predominantly compress visual inputs through feature fusion, token selection, or hybrid strategies. Q-Former~\cite{li2023blip} and Resampler~\cite{bai2023qwen} use trainable queries to aggregate visual context into fixed-length representations, which can then be fused with textual inputs. These architectures significantly reduce token count but may lose fine-grained visual cues, especially in high-resolution settings. FocusLLaVA~\cite{zhu2024focusllava} addresses this by introducing dual expert samplers guided by both visual and textual attention, effectively improving relevance alignment under compression. TokenPacker~\cite{li2024tokenpacker} injects high-resolution visual details into a compressed representation through a dynamic slicing and reassembly scheme. Meanwhile, LLaVA-Mini~\cite{zhangllava} explores extreme compression, reducing image input to a single token using query-based fusion and modality pre-fusion, albeit with potential semantic loss in complex scenes. Other methods such as VisionZip~\cite{yang2024visionzip}, LDPv2~\cite{chu2024mobilevlm}, and Abstractor~\cite{cha2024honeybee} explore various forms of spatial pooling or local detail preservation, but often focus on either visual saliency or computational efficiency alone, without considering instruction priors. In contrast to prior efforts, our work introduces a Top-Down Compression framework that explicitly incorporates instruction-aware global fusion before performing token filtering. 
\section{Methodology}

In this section, we first present the overall framework of LLaVA-Meteor, which generates instruction-following responses conditioned on visual inputs (Section~\ref{sec:meteor}). We then detail its two key components for efficient visual token compression: Flash Global Fusion (Section~\ref{sec:fgf}) and Visual-Native Selection (Section~\ref{sec:vns}). Following the proposed Top-Down Compression, our method enables high-resolution image understanding while significantly reducing memory and computational overhead.

\begin{figure}
  \centering
  \resizebox*{1\linewidth}{!}{\includegraphics[width=\columnwidth]{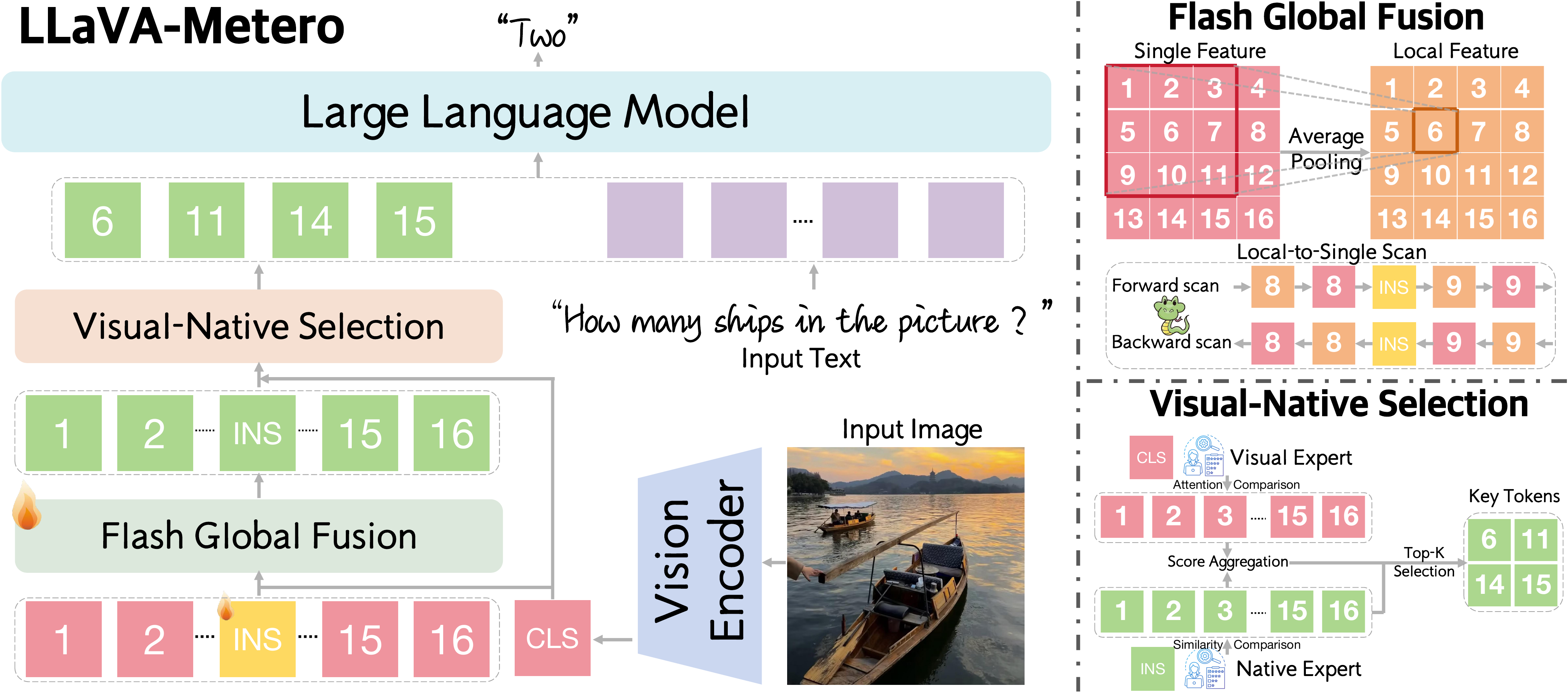} }
  \caption{Overview of the proposed LLaVA-Meteor, which adopts a two-stage Top-Down Compression pipeline to efficiently project visual tokens for instruction-following tasks. An input image is first encoded by a vision encoder into dense visual tokens, including a class token (CLS). We omitted the partitioning step for a clearer presentation. The Flash Global Fusion module (Top) propagates semantic context and instruction-level priors through efficient global interactions, enhanced by a local-to-single scanning strategy that improves local awareness. A learnable instruction token (INS) is inserted to aggregate task-relevant cues during fusion. Subsequently, the Visual-Native Selection module (Down) evaluates token importance from two perspectives: a class-based visual expert and an instruction-based native expert. Their scores are aggregated to select a compact set of key tokens, which are passed to the large language model along with the input to generate the final response.}
  \label{fig:pipeline}

\end{figure}

\subsection{Overview of LLaVA-Meteor}
\label{sec:meteor}
We build our model upon the widely used high-resolution vision-language framework LLaVA-UHD~\cite{guo2024llava}, and the overall architecture is illustrated in Figure~\ref{fig:pipeline}. Given an input image $I \in \mathbb{R}^{H \times W \times \text{3}}$ with arbitrary resolution and aspect ratio, we follow the LLaVA-UHD strategy by first partitioning the image into $N$ variable-sized sub-images $\{I_j\}_{j=\text{0}}^{N}$, where $I_\text{0}$ denotes the resized global image. This partitioning facilitates scalable and efficient encoding of high-resolution content. Each sub-image $I_j$ is processed by a pre-trained vision encoder, CLIP ViT-L/336px~\cite{radford2021learning}, which produces a sequence of visual tokens $\{T_j\} \in \mathbb{R}^{(H_u W_u) \times C}$ along with a corresponding class token ${CLS_j} \in \mathbb{R}^{1 \times C}$. To simultaneously achieve high accuracy and efficiency, we replace the original token compression layer with our proposed Flash Global Fusion and Visual-Native Selection modules. Through the Top-Down Compression, only the Top-$K$ most informative tokens $\{F_j\} \in \mathbb{R}^{K \times D}$, where $K \ll H_t \times W_t$, are retained from each sub-image. Finally, the compressed token sequences $\{F_j\}_{j=\text{0}}^{N}$ are concatenated and combined with the instruction text. This multimodal input is then fed into a large language model, such as Vicuna~\cite{chiang2023vicuna}, to generate the corresponding instruction-following response.

\subsection{Flash Global Fusion}
\label{sec:fgf}
To enrich the representational capacity of visual tokens before compression, we introduce a global context modeling module that enables efficient long-range interaction and instruction-aware feature enhancement. Unlike prior fusion strategies that directly reduce token count at the cost of information loss, our approach retains the full token sequence while refining its contextual semantics, thereby laying a strong foundation for the subsequent selection stage.

\paragraph{Global Context Propagation.} To this end, we employ state space models (SSMs) as the core computational operator due to their ability to model long sequences with linear complexity. In particular, we adopt the advanced selective scan SSMs introduced by Mamba~\cite{mamba} as a lightweight yet effective backbone, enabling fast and global context propagation across the token sequence. This allows each token to incorporate holistic information from other tokens within the same sub-image, while maintaining high throughput on long visual sequences. To compensate for the lack of 2D spatial inductive bias in Mamba, we propose a local-to-single scanning strategy that explicitly enhances local spatial modeling. Specifically, before processing $i^{th}$ token $T^{i}$ (Omit the index of sub-image for clarity
), we first gather its local context by constructing a $\text{3} \times \text{3}$ window centered around the target token in the 2D feature map. This window captures the immediate spatial neighborhood, which is critical for modeling visual patterns. Then, we perform spatial downsampling on these tokens within the window, aggregating them into a single representative token as $TL^{i}$. Therefore, our local-to-single scanning strategy can be succinctly expressed as follows:

\begin{equation}
\begin{aligned}
&forward\ scan : TL^{i-1} \rightarrow T^{i-1} \rightarrow TL^i \rightarrow T^i \rightarrow TL^{i+1} \rightarrow T^{i+1},\\
&backward\ scan : T^{i-1} \leftarrow TL^{i-1} \leftarrow T^i \leftarrow TL^i \leftarrow T^{i+1} \leftarrow TL^{i+1}.
\end{aligned}
\end{equation}

Here, we reverse the scan order within the arrangement of token sequence to capture robust features. This allows us to perform the global context propagation follows:
\begin{equation}
    \{F^{i}\}^{H_u W_u}_{i=\text{1}} = SSMs(L2Sscan(\{TL^{i}\}^{H_u W_u}_{i=\text{1}},\{T^{i}\}^{H_u W_u}_{i=\text{1}})), 
    \label{equ:L2Sscan}
\end{equation}
where $L2Sscan(\cdot)$ denotes the local-to-single and $\{F^{i}\}^{H_u W_u}_{i=1}$ denotes the ouputs tokens.

\paragraph{Instruction Preference Summarization.}Additionally, we insert a shared learnable instruction token, denoted as $INS \in \mathbb{R}^{\text{1} \times C}$, into the center of each sub-image’s token sequence prior to the fusion process. This instruction token is trained to implicitly capture instruction-relevant cues by attending to the full visual context during global context propagation. As a result, it serves as a task-aware prior that guides the token selection process in subsequent Visual-Native Selection stage.

\subsection{Visual-Native Selection}
\label{sec:vns}
Following the enriched token representations produced by the Flash Global Fusion module, we apply a Visual-Native Selection mechanism to perform token compression. This module evaluates the importance of each token from two complementary perspectives: visual significance and instruction relevance, aiming to select the most informative subset for downstream language modeling. By leveraging both a general visual expert and a task-aware native expert, this dual-expert strategy allows the model to retain semantically critical tokens while eliminating redundant ones, thereby significantly improving computational efficiency without compromising task performance.

\paragraph{Visual Expert Scoring.}
To capture general visual importance, we utilize the class token from the frozen vision encoder as a visual expert, as it is known to encode global semantic information. Specifically, we measure the importance of each visual token by computing its attention weight with respect to the class token, serving as a proxy for visual saliency:

\begin{equation}
VS^i_j = \frac{AttnScore(CLS, T^i)}{\sum_{i=\text{1}}^{H_u W_u} AttnScore(CLS, T^i)},
\end{equation}

where $VS^i$ denotes the visual importance score for the $i^{th}$ toke, and $AttnScore(\cdot)$ is derived from the attention map of the frozen vision encoder. Note that the attention distribution includes the self-attention of the class token itself, which causes the total attention over the other tokens to deviate from 1. To address this, we normalize the attention weights across all non-class tokens to form a valid probability distribution. This ensures that the visual expert scores are both interpretable and comparable across different sub-images. By this design, tokens receiving higher scores from the class token are regarded as more visually informative, as they typically correspond to salient objects, prominent regions, or semantically meaningful structures within the image.

\paragraph{Native Expert Scoring.} Complementing the general-purpose visual expert, the native expert is designed to capture instruction preferences. We leverage the instruction token, which is optimized within the Flash Global Fusion module, as a task-aware representation trained to encode high-level semantic intent under vision-language supervision. To evaluate instruction relevance, we compute the similarity between each token and the instruction token using a softmax-normalized dot product:

\begin{equation}
NS^{i} = \frac{e^{\langle F^{i}, INS \rangle}}{\sum_{i=\text{1}}^{H_u W_u} e^{\langle F^{i}, INS \rangle}},
\end{equation}

where $NS^i$ denotes the instruction-aware score of the $i^{th}$ token, and $F^i$ is the token output from the Flash Global Fusion module. To ensure compatibility with the visual expert scores in terms of numerical scale, we apply the softmax operation over all tokens in the same sub-image. This transforms both expert scores into probability-like weights, enabling stable and interpretable fusion in the subsequent selection stage. By leveraging the instruction alignment encoded in instruction token, the model is able to identify semantically important tokens even when they correspond to visually subtle features, thereby overcoming the limitations of saliency-based compression methods.

\paragraph{Score Aggregation and Top-$K$ Selection.}
To integrate the complementary signals from both experts, we compute a fused importance score for each token via weighted aggregation:

\begin{equation}
AS^i = \lambda \cdot VS^i + (1 - \lambda) \cdot NS^i,
\end{equation}

where $\lambda \in [\text{0},\text{1}]$ controls the relative contributions of the visual and native experts. Here, we set $\lambda = \text{0.8}$ as the default value, as it consistently yields favorable performance across multiple datasets. Nevertheless, dataset-specific tuning of $\lambda$ may further improve performance in certain cases. To facilitate stable convergence during early training, we adopt a progressive weighting scheme: the native expert’s contribution is linearly increased from 0 to $\text{1}-\lambda$ over the first few training epochs. Once the aggregated scores $\{AS^i_j\}$ are obtained, we perform token selection over the output sequence from the Flash Global Fusion module, rather than the raw visual tokens. Specifically, we select the Top-$K$ tokens with the highest fused scores:

\begin{equation}
\mathcal{Q}_K = TopK\left( \{ AS^i \}_{i=\text{1}}^{H_u W_u} \right), \quad \{F^i\}_{i \in \mathcal{Q}_K} \subseteq \{F^i\}_{i=1}^{H_u W_u},
\end{equation}

where $\mathcal{Q}_K$ denotes the indices of the selected tokens. This ensures that the retained subset is derived from the enriched sequence, which integrates both global visual context and instruction relevance, thereby maximizing informativeness while maintaining a compact input for the LLMs.

\section{Experiments}
In this section, we present a comprehensive empirical evaluation of LLaVA-Meteor to assess its performance. We begin by detailing the implementation setup (Section~\ref{sec:im} and ~\ref{sec:db}), followed by an analysis of results across nine widely-used benchmarks in comparison with state-of-the-art models (Section~\ref{sec:sota}). Finally, we provide further insights through interpretive analyses (Section~\ref{sec:aba}).

\subsection{Implementation Details}
\label{sec:im}
Our model, LLaVA-Meteor, is constructed based on the widely-used frameworks of LLaVA-1.5~\cite{liu2024improved} and LLaVA-UHD~\cite{guo2024llava}. Specifically, we employpre-trained CLIP ViT-L/336px~\cite{radford2021learning} as the visual encoder, operating at a default resolution of ${\text{336}\times\text{336}}$ to obtain $\text{24} \times \text{24} = \text{576}$ tokens. Language component is powered by Vicuna-13B~\cite{chiang2023vicuna}, and we replace the standard compression layer with our proposed Top-Down Compression module, serving as the projection module linking vision and language. During image slicing, minor adjustments (up to 7–8 pixels) may be made to ensure compatibility with patch-based processing. We follow LLaVA-UHD to slice each input image into $N$ sub-images, the total number of visual tokens passed to the language model becomes $\text{32} \times (N+1)$, including those from a low-resolution global view. We cap $N$ at 6, enabling support for images up to $\text{672}\times\text{1008}$ in resolution. In pretraining phase, only the Top-Down Compression module is trained for one full epoch. We apply the AdamW optimizer with a learning rate of $\text{1} \times \text{10}^{\text{-3}}$ and a cosine decay schedule. The global batch size is set to 256. This phase takes approximately 3 hours on 8 H800 GPUs. For instruction tuning, the visual encoder is kept fixed while both the visual compression and the language model are fine-tuned. The learning rate is reduced to $\text{2} \times \text{10}^{\text{-5}}$, and the batch size is set to 128. Other training configurations remain consistent with the first stage. This stage requires approximately 10 hours on the same hardware.

\subsection{Datasets and Benchmarks}
\label{sec:db}
\textbf{Datasets.} We follow the training datasets as LLaVA-UHD~\cite{guo2024llava}. In the first stage, we utilize CC-595K dataset~\cite{liu2024improved} to finetune the projector. In the second stage, for normal resolution model, we use the dataset consists of a diverse mixture of 656K samples, drawn from LLaVA-Instruct~\cite{liu2023visual}, TextCaps~\cite{sidorov2020textcaps}, GQA~\cite{hudson2019gqa}, OCR-VQA~\cite{mishra2019ocr}, and Visual Genome~\cite{krishna2017visual} for instruction tuning.

\textbf{Benchmarks.} We adopt $12$ popular benchmarks to evaluate our model, including: (1) text-oriented VQA such as TextVQA (VQA$^{T}$)~\cite{singh2019towards}, DocVQA (VQA$^{D}$)~\cite{mathew2021docvqa}, ChartQA (QA$^{C}$)~\cite{masry2022chartqa} and InfoVQA (VQA$^{I}$)~\cite{mathew2022infographicvqa}; (2) general VQA such as GQA~\cite{hudson2019gqa}, VQA$^{v2}$~\cite{goyal2017making} and  VizWiz~\cite{gurari2018vizwiz}; (3) comprehensive evaluation such as MMB
~\cite{liu2024mmbench}, MMVet~\cite{yu2023mm}, MMMU~\cite{yue2024mmmu}, POPE ~\cite{li2023evaluating} and  SEED~\cite{li2023seed}.

\begin{figure}
  \centering
  \resizebox*{1\linewidth}{!}{\includegraphics[width=\columnwidth]{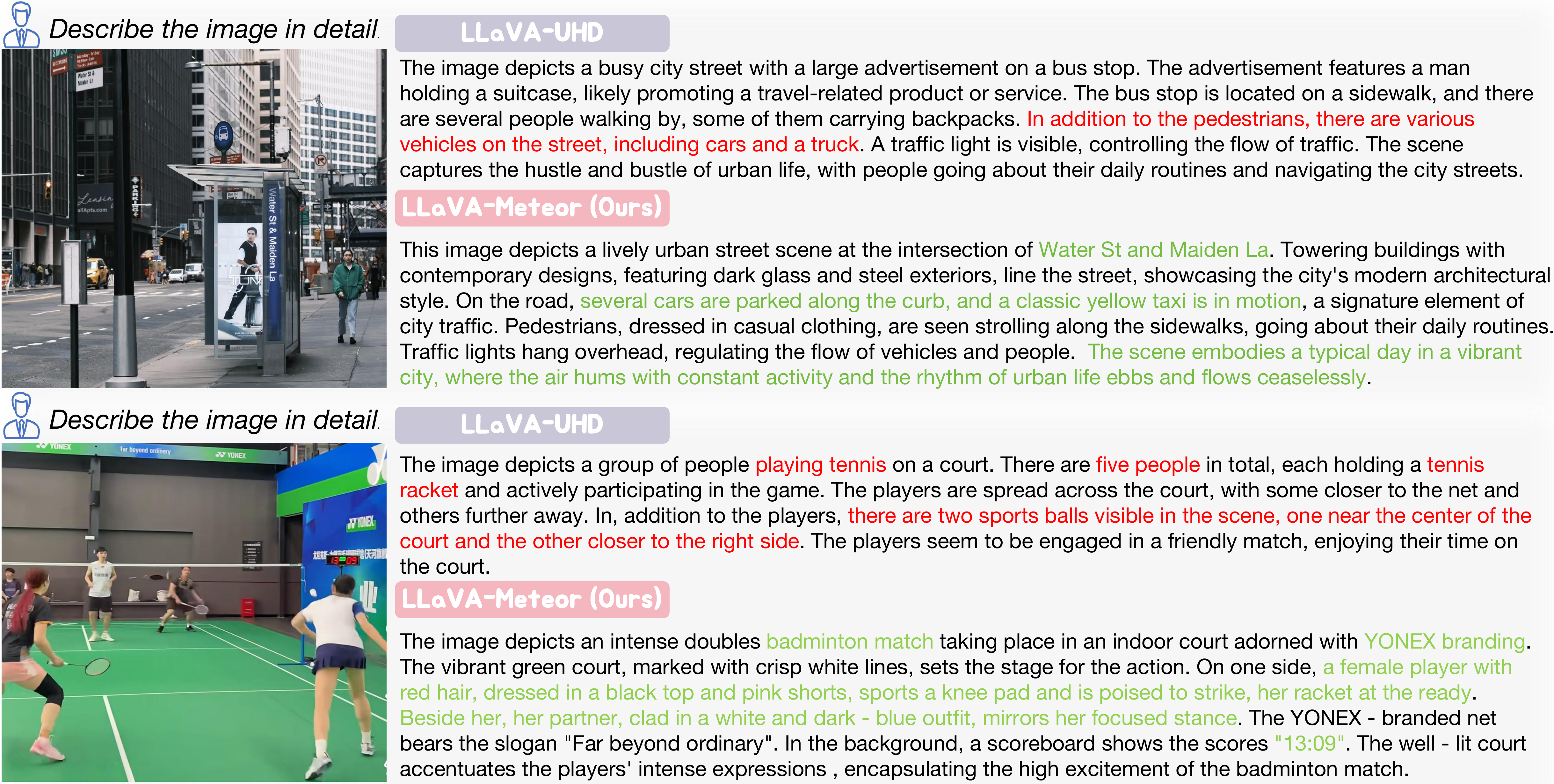} }
  \caption{Qualitative comparison of LLaVA-UHD and our LLaVA-Meteor.}
  \label{fig:vis1}
  
\end{figure}

\begin{figure}
  \centering
  \resizebox*{1\linewidth}{!}{\includegraphics[width=\columnwidth]{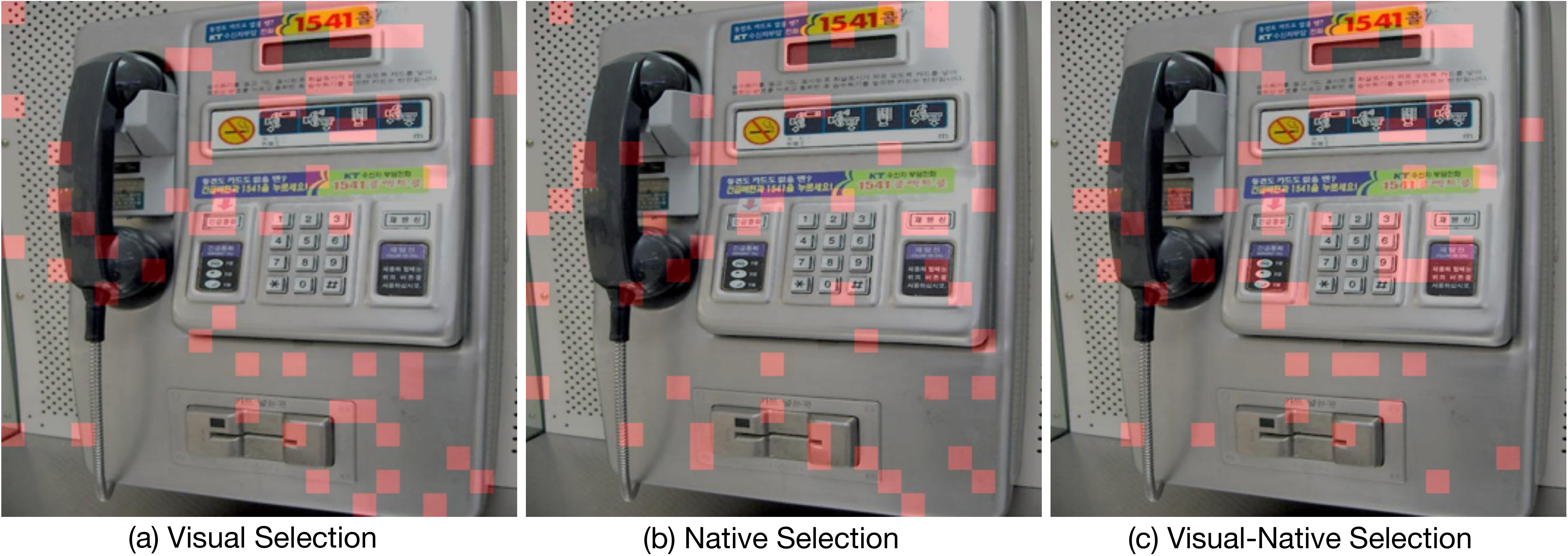} }
  \caption{Qualitative comparisons for different expert selection strategies.}
  \label{fig:vis2}
\end{figure}

\subsection{Compared with SOTA Methods}
\label{sec:sota}
We conduct a comprehensive comparison between our proposed LLaVA-Meteor and prior state-of-the-art vision-language models, with a particular focus on LLaVA-UHD, upon which our work is built. The results in Table~\ref{tab:sota} demonstrate that our model achieves competitive or superior performance across diverse benchmarks while utilizing significantly fewer visual tokens. Since images may be divided into different numbers of sub-images depending on their resolution and aspect ratio, we report the average number of visual tokens per image across datasets in the table. For clarity and fair comparison, we refer to the token count per sub-image when describing and analyzing model behavior throughout the text. Specifically, when using the same token budget as LLaVA-UHD (144 tokens for on sub-images), LLaVA-Meteor yields consistent improvements on almost all tasks, including 69.9 on VQA$^T$, 39.2 on VAQ$^I$, 64.9 on GQA, 82.4 on VQA$^{v2}$ and 89.9 on POPE. These gains indicate enhanced comprehension in both text-oriented and reasoning-heavy vision-language scenarios. Notably, the average performance increases by +2.0 points, validating the overall robustness of our approach. More impressively, when we reduce the token count to 64 and 32—representing 20\% and 5\% of the original token length for on sub-images—LLaVA-Meteor still maintains strong performance. For instance, with only 64 tokens, our model still achieves 58.6 on QA$^C$ and 66.9 on SEED, with only a marginal drop of -0.4 and -0.8 compared to the 144-token setting, while significantly improving efficiency. Compared to LLaVA-UHD, even our most compressed 32-token variant retains competitive accuracy while reducing token usage by over 95\%. This highlights the effectiveness of our Top-Down Compression framework, which selects a semantically rich token subset without sacrificing representational fidelity. To provide a more intuitive illustration of the effectiveness of LLaVA-Meteor, we further visualize the input image, instruction, and the corresponding response, as shown in Figure~\ref{fig:vis1}. Overall, our method demonstrates a compelling balance between performance and computational cost, making it highly suitable for resource-constrained scenarios and real-world deployments.

\begin{table*}
\centering
\caption{Comparison of recent vision-language models across text-oriented VQA, general VQA, and comprehensive benchmarks. }
\label{tab:sota}
\resizebox*{1\linewidth}{!}{
\begin{tabular}{lll|c|cccccccccccc|c}
\toprule
Model & LLM & PT/IT & Token & VQA$^{T}$ & VQA$^{D}$ & QA$^{C}$ & VQA$^{I}$ & GQA & VQA$^{v2}$ & VizWiz & MMB &MMVet & MMMU & POPE & SEED & Avg \\
\midrule
MobileVLM V2~\cite{chu2024mobilevlm} & Mobilellama-2.7B &1.2M/3.6M &144 &52.1& -& -& -& 59.3& - &-& -& - &-& 84.3& -& -\\
BLIP-2~\cite{li2023blip}& Vicuna-13B &129M/- & 32& 42.5& -& -& - &41.0& 65.0 &19.6& -& - &-& 85.3 &49.7& -\\
Insturct-BLIP~\cite{panagopoulou2023x}& Vicuna-7B &129M/1.2M& 64 &50.1& - &- &-& 49.5& - &34.5& -& 26.3 &-& -& -& -\\
QwenVL~\cite{bai2023qwen} & Qwen-7B& 1.4B/50M& 256& 63.8& 65.1& \underline{65.7}& - &59.3& 78.8& 35.2& -& -& -& -& 62.3& -\\
VILA~\cite{lin2024vila}& Llama2-7B& 50M/1M& 576& 64.4 &- &58.6& - &62.3& 79.9& 57.8& 68.9 & 34.9& -& 85.5& -& -\\
MobileVLM V2~\cite{chu2024mobilevlm}& Vicuna-7B& 1.2M/3.6M& 144& 62.3& -& -& -& 62.6& - &- &- &- &-& 85.3& -& -\\
Mini-Gemini~\cite{li2024mini}& Vicuna-7B& 1.2M/1.5M& 576& 65.9& -& -& -& -& -& -& 68.5& \underline{46.0}& \underline{38.1} & -& -& -\\
LLaVA-1.5~\cite{liu2024improved} & Vicuna-7B& 558K/665K& 576& 58.2& 28.1& -& 25.8& 63.3& 78.5& 50.0& 64.3& 31.1& 35.3& 85.9& 66.1& -\\
TokenPacker~\cite{li2024tokenpacker}& Vicuna-7B& 558K/665K& 144& -& -& -& -& 61.9& 77.9& 52.0& 65.1& 33.0& - &87.0& -& -\\
InternVL2~\cite{chen2024far}& Internlm2.5-7B& 558K/665K& 256& 49.7& 26.9& 18.1& 21.8& 63.0& 77.8& 50.6& 70.9& 34.1& 39.2& 86.8& \underline{71.1}& 50.8\\
\midrule
\rowcolor{gray!15}\multicolumn{17}{c}{\textit{High-resolution LLMs}}\\
\midrule
Monkey~\cite{li2024monkey}& Qwen-7B& -/1.44M& \textasciitilde 1024& 67.7& \underline{66.5} & 36.1& -& 60.7& 80.3& \textbf{61.2}& -& -& -& -& -& -\\
TokenPacker-HD~\cite{li2024tokenpacker}& Vicuna-7B& 1.2M/1.5M& \textasciitilde 954& 68.0& 60.2& -& -& -& 81.2& 54.7& 67.4& -& 35.4& -& -& -\\
Mini-Gemini-HD~\cite{li2024mini}& Vicuna-7B& 1.2M/1.5M& 2880& \underline{68.4}& 65.0 & -& -& -& 80.3& 54.6& 65.8& 41.3& 36.8& 86.8& -& -\\
FastViTHD~\cite{vasu2024fastvlm} &Qwen2-7B & 558K/1.1M& 256 & 64.4 &- & -& -& -& 63.1 & - & - & -& -& 88.1 &- & -\\
LLaVA-UHD~\cite{guo2024llava}& Vicuna-13B& 595K/665K& \textasciitilde 256& 67.7&62.6&56.3&36.8&63.8&81.7&56.1&68.0&42.1&35.5&\underline{89.1}&65.6&60.4\\
LLaVA-NeXT~\cite{liu2024llavanext} &Vicuna-7B& 558K/765K& \textasciitilde 2880& 64.9& 74.4& 54.8& \underline{37.1}& \underline{64.2} & \underline{81.8}& 57.6& 68.1& 43.9& 35.8& 86.5& 68.2& 61.4\\
InternVL2-HD~\cite{chen2024far}& Internlm2.5-7B& 558K/770K& \textasciitilde 1282& 65.6& \textbf{72.6}& \textbf{69.8}& 30.9& 63.2& 78.9& 56.3& \textbf{72.1}& 35.7& \textbf{39.9}& 87.3& \textbf{73.4}& \underline{62.1}\\
\midrule
\rowcolor{gray!15}\multicolumn{17}{c}{\textit{Ours}}\\
\midrule
LLaVA-Meteor & Vicuna-13B & 595K/665K & \textasciitilde 256 & \textbf{69.9} & 64.2 & 59.0 & \textbf{39.2} & \textbf{64.9} & \textbf{82.4} & \underline{59.3} & \underline{69.4} & \underline{44.7} & 37.5 & \textbf{89.9} & 67.7&\textbf{62.4}\\
\multicolumn{3}{c|}{\textit{{\textcolor{gray}{ compare to LLaVA-UHD}} }} &\textcolor{red}{100\%}& \textcolor{red}{+2.2}& \textcolor{red}{+1.6} & \textcolor{red}{+2.7} & \textcolor{red}{+2.4} & \textcolor{red}{+1.1} & \textcolor{red}{+0.7} & \textcolor{red}{+3.2} & \textcolor{red}{+1.4} & \textcolor{red}{+2.6} & \textcolor{red}{+2.0} & \textcolor{red}{+0.8} & \textcolor{red}{+2.1}&\textcolor{red}{+2.0}\\
\midrule
LLaVA-Meteor & Vicuna-13B & 595K/665K & \textasciitilde 114 & 68.3 &63.1 & 58.6 & 37.7 & 64.6 & 81.8 & 57.1 & 68.4 & 42.7 & 34.6 & 88.7 & 66.9 &  61.0\\
\multicolumn{3}{c|}{\textit{{\textcolor{gray}{ compare to LLaVA-UHD}} }} &\textcolor{red}{44.5\%}& \textcolor{red}{+0.6}& \textcolor{red}{+0.5} & \textcolor{red}{+2.3} & \textcolor{red}{+0.9} & \textcolor{red}{+0.8} & \textcolor{red}{+0.1} & \textcolor{red}{+1.0} & \textcolor{red}{+0.4} & \textcolor{red}{+0.6} & \textcolor{blue}{-0.8} & \textcolor{blue}{-0.5} & \textcolor{red}{+1.3}&\textcolor{red}{+0.6}\\
\midrule 
LLaVA-Meteor & Vicuna-13B & 595K/665K & \textasciitilde 56 & 65.0 & 58.4 & 56.5 & 37.1 & 62.4 & 81.2 & 55.3 & 68.0 & 41.6 & 34.2 & 87.2 & 64.8 & 59.3\\
\multicolumn{3}{c|}{\textit{{\textcolor{gray}{ compare to LLaVA-UHD}} }} &\textcolor{red}{21.8\%}& \textcolor{blue}{-2.7}& \textcolor{blue}{-4.2} & \textcolor{red}{+0.2} & \textcolor{red}{+0.3} & \textcolor{blue}{-1.4} & \textcolor{blue}{-0.5} & \textcolor{blue}{-0.8} & \textcolor{red}{+0.0} & \textcolor{blue}{-0.5} & \textcolor{blue}{-1.3} & \textcolor{blue}{-1.9} & \textcolor{blue}{-0.8}&\textcolor{blue}{-1.1}\\
\bottomrule
\end{tabular}
}

\end{table*}

\subsection{Ablation Study}
\label{sec:aba}

\paragraph{Scanning Strategies.} We conduct an ablation study to assess the impact of different scanning strategies within the Flash Global Fusion module. Specifically, we compare three configurations: (\textit{$\romannumeral1$}) the single scan strategy, which processes tokens sequentially without local context modeling; (\textit{$\romannumeral2$}) the local-to-single scan strategy, which enhances each token representation by aggregating its $\text{3} \times \text{3}$ spatial neighborhood before propagation. Experimental results in Table~\ref{tab:aba1} demonstrate that the local-to-single scan strategy consistently achieves better performance than the single scan strategy. This highlights the importance of integrating local spatial dependencies, especially in scenarios requiring fine-grained visual understanding. Furthermore, 
we also attempt to combined these two strategy but find it only provides marginal improvements over the local-to-single, the gains are not significant and come at the cost of 50\% additional computation in compression modules. In light of these findings, we adopt the local-to-single scan strategy as the default configuration in all experiments, as it offers a favorable balance between accuracy and efficiency.  Additionally, we recommend that when operating on high-resolution feature maps, the local window size can be further increased (\textit{e.g.}, to $\text{5} \times \text{5}$ or $\text{7} \times \text{7}$) to enable richer multi-scale context modeling and improve the perception of spatial hierarchies.

\paragraph{Expert Selection Strategies.}To better understand the roles of different expert signals in our Visual-Native Selection module, we conduct ablation studies comparing three configurations for Top-$K$ token selection: (\textit{$\romannumeral1$}) using only the visual expert, (\textit{$\romannumeral2$}) using only the native expert, and (\textit{$\romannumeral3$}) using both experts jointly via weighted aggregation. Experimental results in Table~\ref{tab:aba2} show that for tasks requiring dense prediction, such as object understanding, the performance of using only the visual expert is comparable to that of the combined visual-native expert. This suggests that the visual expert, which originates from a pre-trained vision encoder, is sufficient to capture general semantic structure in these scenarios. However, in more challenging tasks that involve reasoning and attribute discrimination, the visual-native configuration demonstrates a clear advantage. The native expert contributes instruction-aware cues that enhance the model's ability to focus on contextually important but visually less prominent tokens, leading more accurate alignment with task objectives. In contrast, relying solely on the native expert results in inferior performance. This approach requires the model to learn task-specific knowledge entirely from vision-language supervision without support from visual encoder. Such reliance not only reduces overall accuracy but also leads to unstable training behavior, especially in the early training stages, as indicated by more frequent and larger fluctuations in the loss values. To facilitate a clearer understanding of the patch selection behavior of each expert, we present detailed visualizations in Figure~\ref{fig:vis2}. Based on these observations, we recommend using both experts in combination. The weighting parameter $\lambda$ between the two experts can be adjusted to suit the nature of the task. For example, lower values of $\lambda$ are preferable in instruction-focused domains, while higher values can be beneficial when visual saliency dominates. In our default configuration, we set $\lambda = \text{0.8}$, as it yields consistently strong results across a variety of benchmarks.

\begin{table*}
\centering
\caption{Comparison of  different scanning strategies.}
\label{tab:aba1}
\resizebox*{1\linewidth}{!}{
\begin{tabular}{c|cccccccccccc|c}
\toprule
Scan Strategy & VQA$^{T}$ & VQA$^{D}$ & QA$^{C}$ & VQA$^I$ & GQA & VQA$^{v2}$ & VizWiz & MMB &MMVet & MMMU & POPE & SEED & Avg \\
\midrule
Single &68.9 &63.0& 58.1 & 38.4 & 64.0 & 81.8 & 57.5 & 68.1 & 42.0 & \textbf{37.5} & 88.7 & 65.4 &61.1\\
Local-to-Single & \textbf{69.9} & \textbf{64.2} & \textbf{59.0} & \textbf{39.2} & \textbf{64.9} & \textbf{82.4} & \textbf{59.3} & \textbf{69.4} & \textbf{44.7} & \textbf{37.5} & \textbf{89.9} & \textbf{67.7}&\textbf{62.4} \\
\bottomrule
\end{tabular}}
\end{table*}

\begin{table*}
\centering
\caption{Comparison of different expert selection strategies.}
\centering
\label{tab:aba2}
\resizebox*{1\linewidth}{!}{
\begin{tabular}{c|cccccccccccc|c}
\toprule
Expert Selection Strategy & VQA$^{T}$ & VQA$^{D}$ & QA$^{C}$ & VQA$^I$ & GQA & VQA$^{v2}$ & VizWiz & MMB &MMVet & MMMU & POPE & SEED & Avg \\
\midrule
Visual & 69.7 & 63.5 & \textbf{59.3} & 37.8 & 63.8 & 82.1 & 57.2 & 66.9  & 41.7 & \textbf{38.1} & 88.0 & 65.5 &61.1\\
Native & 67.1 & 62.3 & 57.7 & \textbf{37.9} & 61.9 & 79.4 & 56.6 & 66.2 & 40.0 & 31.1 & 86.8 & 63.4 & 59.2\\
Visual-Native & \textbf{69.9} & \textbf{64.2} & 59.0 & 37.8 & \textbf{64.9} & \textbf{82.4} & \textbf{59.3} & \textbf{69.4} & \textbf{44.7} & 37.5 & \textbf{89.9} & \textbf{67.7}&\textbf{62.4}\\
\bottomrule
\end{tabular}}
\end{table*}

\begin{wraptable}{r}{0.53\linewidth}
	\centering
    \vspace{-0.5cm}
	\setlength{\abovecaptionskip}{0cm}
    \captionsetup{width=.53\textwidth} 
\caption{Efficiency comparison with different model.}
\label{tab:aba3}
{
\resizebox{0.53\textwidth}{!}{
\begin{tabular}{cc|cccc}
\toprule
Model & Tokens & Parameters & Train & TPS & Avg\\
\midrule
LLaVA-UHD~\cite{guo2024llava} & 144 & 137.84M  & 17.2h & 28.4 &60.4\\
\midrule
LLaVA-Meteor & 144 & \textbf{37.08M} & 13.7h & 29.1&\textbf{62.4}\\
LLaVA-Meteor & 64 & \textbf{37.08M}  & 12.9h & 31.3&61.0\\
LLaVA-Meteor & 32 & \textbf{37.08M}& \textbf{12.1h} & \textbf{32.9}&59.3\\
\bottomrule
\end{tabular}}}
\vspace{-0.5cm}
\end{wraptable}

\paragraph{Model Efficiency.} To validate the efficiency of our proposed framework, we conduct a comprehensive ablation study across four three aspects: model size (parameter count), training time, and TPS (token per second). Note that we only present the parameters of the projector here. We compare our method against sota LLaVA-UHD under identical hardware settings and input resolutions. Experimental results in Table~\ref{tab:aba3} demonstrate that our model consistently achieves favorable efficiency across all metrics. In particular, the overall parameter remains competitive due to the lightweight design of the Flash Global Fusion and training-free Visual-Native Selection modules. More notably, the computational cost is significantly reduced as a result of effective token compression. Since fewer visual tokens are forwarded to the language model, both training and inference stages benefit from lower memory usage and shorter runtime. As the number of retained tokens decreases, we observe the increase in TPS, confirming that token compression directly translates to computational savings. This property makes our model highly suitable for deployment in resource-constrained scenarios or applications requiring real-time response. Overall, the proposed architecture provides a balanced solution that preserves task performance while substantially improving computational efficiency.

\section{Conclusion}
In this paper, we revisit the challenge of efficient vision token projection in the context of visual instruction tuning and introduce a novel compression framework termed Top-Down Compression. By explicitly integrating global fusion with expert-guided token selection, our method effectively bridges the gap between performance and efficiency. Specifically, the proposed Flash Global Fusion module enables lightweight yet holistic context propagation, while the Visual-Native Selection strategy jointly evaluates token significance from both semantic and instruction-aware perspectives. Built upon this paradigm, our model LLaVA-Meteor achieves state-of-the-art or competitive performance across a wide range of vision-language benchmarks, with up to 75\%–95\% reduction in visual tokens. These results demonstrate that our approach not only preserves semantic fidelity but also significantly enhances computational efficiency, making it well-suited for real-world, resource-constrained applications. We hope this work offers a new perspective on balancing compression and task alignment for future vision-language modeling.

\bibliographystyle{plain}

\bibliography{refs.bib}

%%%%%%%%%%%%%%%%%%%%%%%%%%%%%%%%%%%%%%%%%%%%%%%%%%%%%%%%%%%%
%%%%%%%%%%%%%%%%%%%%%%%%%%%%%%%%%%%%%%%%%%%%%%%%%%%%%%%%%%%%

\end{document}